\documentclass[10pt,twocolumn,letterpaper]{article}

\usepackage[pagenumbers]{cvpr}

\usepackage{colortbl}
\usepackage{booktabs,multirow}
\usepackage{makecell}
\usepackage{tabulary}
\usepackage{fontawesome5}

\newlength\savewidth\newcommand\shline{\noalign{\global\savewidth\arrayrulewidth
  \global\arrayrulewidth 1pt}\hline\noalign{\global\arrayrulewidth\savewidth}}

\usepackage{capt-of}
\usepackage[dvipsnames]{xcolor}
\usepackage{xspace}
\usepackage{enumitem}
\usepackage{amsmath,amssymb,amsbsy,amsfonts,dsfont,pifont,bm,bbm,mathrsfs,mathtools,nicefrac}
\usepackage{algorithm,algpseudocode,listings}
\usepackage{booktabs,multirow,adjustbox,diagbox,threeparttable,tabularray}

\newcommand{\method}{\texttt{Ranni}\xspace}

\newcommand{\tocite}[1]{{\color{red} [TO CITE]}}

\definecolor{CQColor}{rgb}{0.0,0.0,1.0} 

\definecolor{CQRColor}{rgb}{1.0,0.0,1.0} 

\newcommand{\myparagraph}[1]{\vspace{3.0pt}\noindent{\bf #1}}

\definecolor{cvprblue}{rgb}{0.21,0.49,0.74}
\usepackage[pagebackref,breaklinks,colorlinks,citecolor=cvprblue]{hyperref}
\usepackage{wrapfig}
\usepackage[capitalize]{cleveref}  
\crefname{section}{Sec.}{Secs.}
\Crefname{section}{Section}{Sections}
\crefname{table}{Tab.}{Tabs.}
\Crefname{table}{Table}{Tables}
\crefname{figure}{Fig.}{Figs.}
\Crefname{figure}{Figure}{Figures}
\crefname{equation}{Eq.}{Eqs.}
\Crefname{equation}{Equation}{Equations}
\title{Ranni: Taming Text-to-Image Diffusion for Accurate Instruction Following}

\author{
    Yutong Feng$^{1}$,
    Biao Gong$^{1}$,
    Di Chen$^{1}$,
    Yujun Shen$^{2}$,
    Yu Liu$^{1}$,
    Jingren Zhou$^{1}$ \\
    {$^1$Alibaba Group\ \ $^2$Ant Group}\\[-2pt]
    {\tt\small \{fengyutong.fyt, a.biao.gong, deechan1994, shenyujun0302\}@gmail.com}\\[-3pt]
    {\tt\small \{ly103369, jingren.zhou\}@alibaba-inc.com}\\  
}

\begin{document}

\twocolumn[{
\maketitle
\begin{center}
    \vspace{-12pt}
    \includegraphics[width=1.0\linewidth]{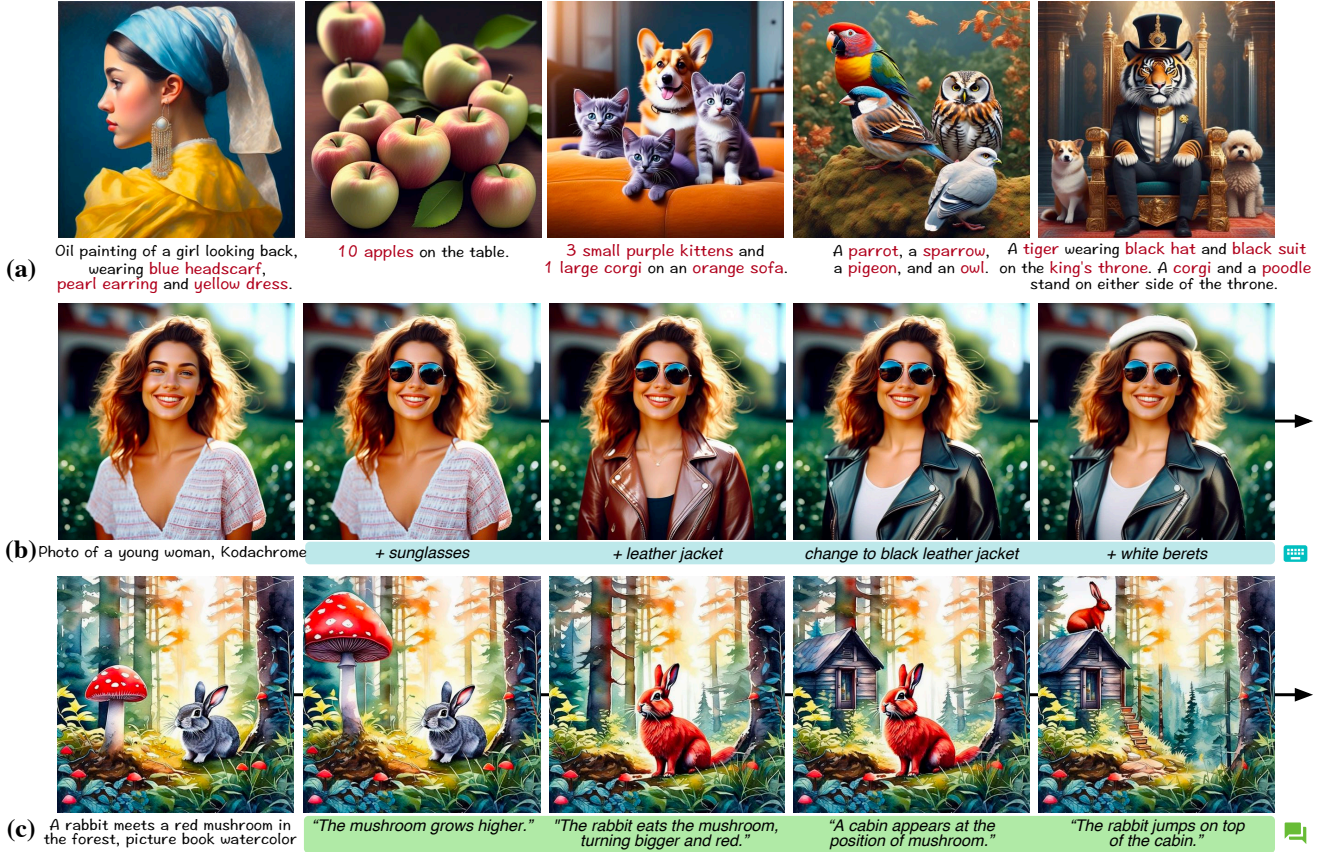}
    \vspace{-18pt}
    \captionof{figure}{
        Samples generated by \method with different interaction manners,  including \textbf{(a) direct generation} with accurate prompt following,  \textbf{(b) continuous generation} with progressive refinement, and \textbf{(c) chatting-based generation} with text instructions.
    }
    \label{fig:teaser}
\end{center}
}]

\begin{abstract} 

\vspace{-3mm}

Existing text-to-image (T2I) diffusion models usually struggle in interpreting complex prompts, especially those with quantity, object-attribute binding, and multi-subject descriptions.
In this work, we introduce a \textbf{semantic panel} as the middleware in decoding texts to images, supporting the generator to better follow instructions.
The panel is obtained through arranging the visual concepts parsed from the input text by the aid of large language models, and then injected into the denoising network as a detailed control signal to complement the text condition.
To facilitate text-to-panel learning, we come up with a carefully designed semantic formatting protocol, accompanied by a fully-automatic data preparation pipeline.
Thanks to such a design, our approach, which we call \method, manages to enhance a pre-trained T2I generator regarding its textual controllability.
More importantly, the introduction of the generative middleware brings a more convenient form of interaction (ie, directly adjusting the elements in the panel or using language instructions) and further allows users to finely customize their generation, based on which we develop a practical system and showcase its potential in continuous generation and chatting-based editing.
The project page is \url{https://ranni-t2i.github.io/Ranni/}.
\vspace{-5mm}

\end{abstract}
    
\section{Introduction}
\label{sec:intro}

Language is the most straightforward way for us to convey perspectives and creativity.
When we aim to bring a scene from  imagination into reality, the first choice is through language description.
This forms the  philosophical basis of text-to-image (T2I) synthesis.
With recent advancements in diffusion models, T2I synthesis demonstrates promising results in terms of high fidelity and diversity~\cite{ho2020denoising,Song2020DenoisingDI,dhariwal2021diffusion,nichol2021improved,ramesh2021zero,ramesh2022,stablediffusion2}.
However, the expressive power of language is also limited, compared with structured, pixel-based image in more diverse distribution.
This hinders the T2I synthesis to faithfully translate a textual description into a precisely corresponding image.
Therefore, current models encounter issues when generating for complex prompts~\cite{t2icompbench}, such as determining the quantity of objects, attribute binding, spatial relationship, and multi-subject descriptions.

For professional painters and designers, they express an imagined scene into a tangible form with a broader range of tools beyond just language, \textit{e.g.,} cascading style sheets (CSS) and designing softwares.
These tools allow for accurate and enriched expression of visual objects, from the perspectives of spatial positions, sizes, relationships, styles, \etc.
By getting closer to the image modality, they achieve more accurate expression and easier manipulation.

In this paper, our goal is to introduce a new image generation approach, which offers the convenience of text-to-image methods, while also providing accurate expression and enriched manipulation capabilities similar to professional tools.
To this end, we present \textbf{\method}, an improved T2I generation framework which translates natural language into a middleware with the help of large language models (LLMs). 
The middleware, which we call \textbf{semantic panel}, acts as a bridge between text and images. 
It provides accurate understanding of text descriptions and enables intuitive image editing.
The semantic panel comprises all the visual concepts that appear in the image. 
Each concept represents a structured expression of an object.
We describe it using various attributes, such as its bounding box, colors, keypoints, and the corresponding textual description.

By introducing the semantic panel, we relax the text-to-image generation with two sub-tasks: \textit{text-to-panel} and \textit{panel-to-image}.
During the text-to-panel, text descriptions are parsed into visual concepts by LLMs, which are gathered and arranged inside the semantic panel.
The panel-to-image process then encodes the panel as a control signal, guiding the diffusion models to capture the details of each concept.
To support efficient training on the above tasks, we present an automatic data preparation pipeline.
It extends the text-image pairs of existing datasets by extracting visual concepts using a collection of recognition models.

Based on the semantic panel, \method also offers a more intuitive way to further edit the generated image. 
Existing diffusion-based methods~\cite{glide,blended_diff,text2live,instruct_p2p} implicitly understand the editing intention through modified prompts or text instructions. 
In contrast, we explicitly map the editing intention to an update of the semantic panel. 
With the rich attributes of visual concepts, we are able to incorporate most editing instructions by composing six unit operations: \textit{addition, removal, resizing, re-position, replacement,} and \textit{attribute revision}. 
The update of the semantic panel can be done manually through a user interface or automatically by LLMs. 
In practical, we study the adaption of advanced LLMs on this task.
The results demonstrate the potential of a fully-automatic \textbf{chatting-based editing} approach, which allows for continuous generation all with text instructions.

\section{Related Work}
\label{sec:related_work}

\noindent\textbf{Text-to-Image Generation Models.}
Diffusion models \cite{ho2020denoising,dhariwal2021diffusion,Song2020DenoisingDI,nichol2021improved,rombach2022high} have become more popular recently compared to GANs \cite{xu2018attngan,zhang2021cross,zhu2019dm} and auto-regressive models \cite{Ding2021CogViewMT,Yu2022ScalingAM,Esser2020TamingTF,Yu2021VectorquantizedIM,Ramesh2021ZeroShotTG,ramesh2021zero} due to their ability to produce high-quality and diverse outputs.
Recent advancements such as Stable Diffusion \cite{stablediffusion2}, UnCLIP \cite{ramesh2022}, and Imagen \cite{saharia2022photorealistic} have demonstrated significant improvements in generating text-to-image with an impressive level of photo-realism.
\method builds upon diffusion model, taming it for better instruction following while maintaining the generation quality.

\noindent\textbf{Controllable Generation beyond Texts.}
Recent works also extend the controllability of diffusion models
by including extra conditions such as inpainting masks \cite{xie2023smartbrush,lhhuang2023composer}, sketches \cite{voynov2023sketch}, keypoints \cite{li2023gligen}, depth maps \cite{stablediffusion2}, segmentation maps \cite{Wang2022PretrainingIA,Couairon2022DiffEditDS}, layouts \cite{rombach2022high}, \etc.
Accordingly, models are modified by incorporating additional encoders through either fine-tuning (\eg, ControlNet \cite{zhang2023adding}, T2I-Adapter \cite{mou2023t2i}) or training from scratch (\eg, Composer \cite{lhhuang2023composer}).

\noindent\textbf{LLM-assisted Image Generation.} 
The large language models (LLMs) have revolutionized various NLP tasks with exceptional generalization abilities, which are also leveraged to assist in text-to-image generation.
LLM-grounded Diffusion~\cite{lian2023llmgroundeddiffusion} and VPGen~\cite{cho2023vpgen} utilize LLMs to infer object locations from text prompts using carefully designed system prompts.
LayoutGPT~\cite{feng2023layoutgpt} improves upon this framework by providing the LLM with retrieved exemplars.
\method further incorporates a comprehensive semantic panel with multiple attributes.
It fully leverages the planning ability of LLMs for accurately following painting and editing tasks.

\begin{figure*}   
  \centering
   \includegraphics[width=0.95\linewidth]{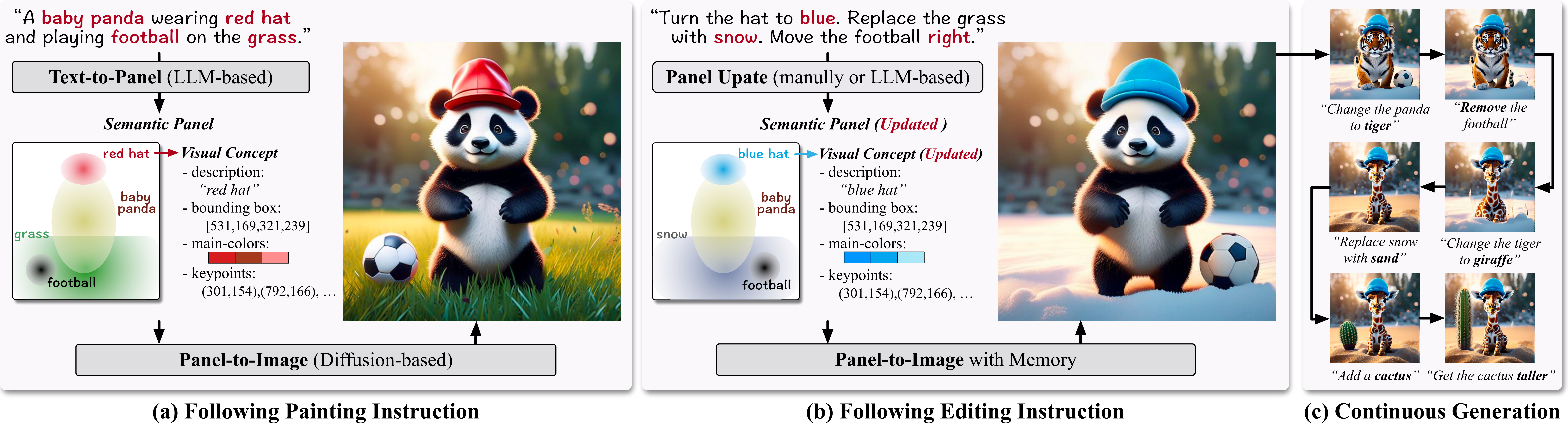}
    \vspace{0mm}
   \caption{The 
   \textbf{framework} of \method for following painting and editing instructions in a \textit{sequential workflow} based on the semantic panel. \textbf{(a)} The painting task is divided into an LLM-assisted text-to-panel, and a diffusion-based panel-to-image generation. \textbf{(b)} The editing task is conducted via the update of previous semantic panel. \textbf{(c)} The image can be further refined with multi-round compounded editing.
   }
       \vspace{-2mm}
   \label{fig:pipeline}
\end{figure*}


\section{Methodology}
\label{sec:method}

We begin by presenting the framework of \method, which utilizes a semantic panel for accurate text-to-image generation.
Next, we expand upon the framework to enable interactive editing and continuous generation.
The entire framework is depicted in \cref{fig:pipeline}.
Lastly, we introduce an automatic data preparation pipeline and the created dataset, which enables the efficient training of \method.

\subsection{Bridging Text and Image with Semantic Panel}

We define the semantic panel as a workspace for manipulating all visual concepts in an image. 
Each visual concept represents an object, and includes its visually accessible attributes (\textit{e.g.,} position and colors).
The semantic panel acts as a middleware between text and image, presenting a structured modeling for text, and a compressed modeling for image. 
By incorporating the panel, we alleviate the pressure of directly mapping text to image.
We include the following attributes for each concept:
1) \textit{text description} for semantic information, 2) \textit{bounding box} for position and size, 3) \textit{main colors} for style, and 4) \textit{keypoints} for shape.
The text-to-image generation is then naturally divided two sub-tasks: \textit{text-to-panel} and \textit{panel-to-image}.

\myparagraph{Text-to-Panel} requires the ability to understand prompts and have a rich knowledge of visual content. 
We adapt the LLM for this task due to its strong performance as a prompt reader and a task-planner. 
We design system prompts to request the LLM for imagining the visual concepts corresponding to the input text. 
When generating multiple attributes of concepts, inspired by chain-of-thought~\cite{chain_of_thought}, we conduct it in a sequential manner.
The whole set of objects is firstly generated with the text descriptions.
Detailed attributes, \textit{e.g.,} bounding boxes, are then generated and arranged towards each object.
The design of chat templates and examples of full conversations are available in the Supplement Material.
Thanks to the zero-shot ability of LLMs, they can generate detailed semantic panels with correct output format.
Furthermore, we enhance the performance of LLM by fine-tuning it to better comprehend visual concepts, especially for more detailed attributes like colors.
This is achieved by utilizing a large dataset consisting of image-text-panel triples.
The details of dataset construction are explained in \cref{chap:dataset}.


\begin{figure*}    
  \centering
   \includegraphics[width=0.95\linewidth]{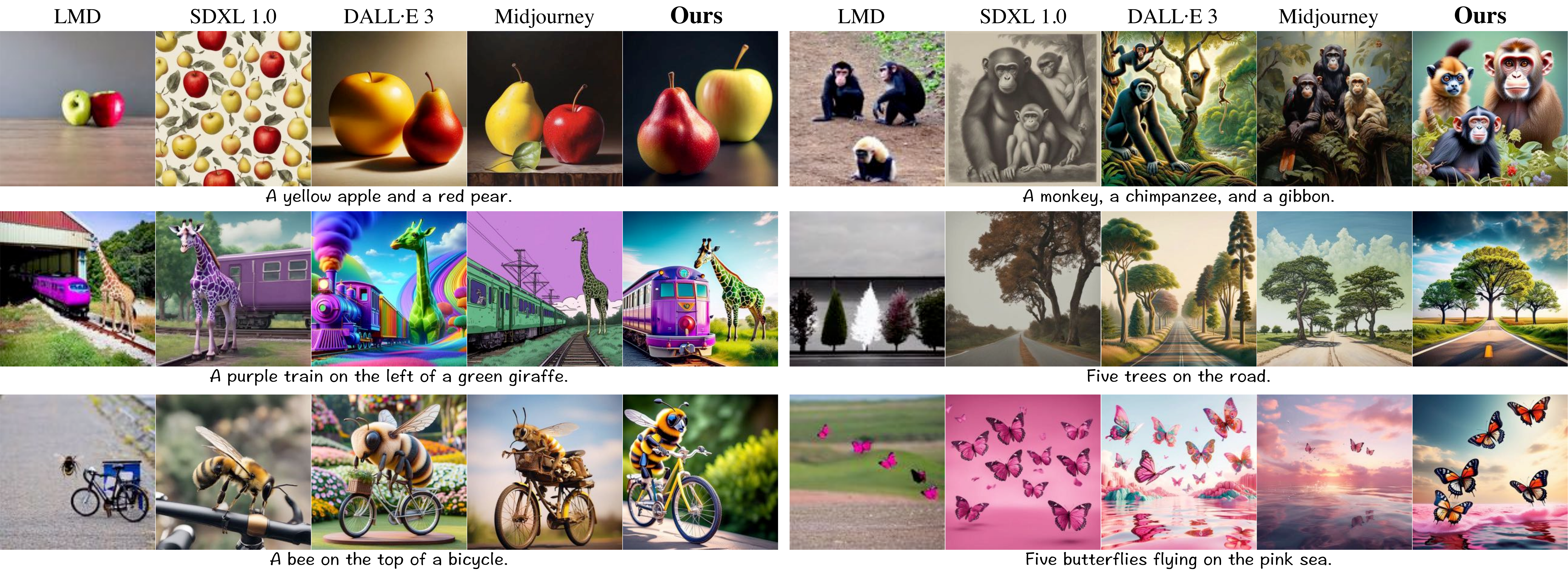}
    \vspace{0mm}
   \caption{
   \textbf{Comparison on text-to-image generation} between \method and representative methods.
   }
   \vspace{-4mm}
   \label{fig:gen_comp}
\end{figure*}

\myparagraph{Panel-to-Image} is a task focused on conditional image generation. 
We implement it using the latent diffusion model~\cite{rombach2022high} as the backbone. 
To begin, all visual concepts within the semantic panel are encoded into a condition map that has the same shape as the image latent.
The encodings of different attributes are as follows:
\setlist[itemize]{leftmargin=15pt}
\begin{itemize}
    \vspace{-7pt}
    \setlength{\itemsep}{0pt}
    \setlength{\parsep}{0pt}
    \setlength{\parskip}{0pt}
    \item \textit{text description}: CLIP text embedding.
    \item \textit{bounding box}: A binary mask with 1s inside the box.
     \item \textit{colors}: Indexed learnable embeddings.
    \item \textit{keypoints}: A binary heatmap with 1s on the keypoints.
    \vspace{-7pt}
\end{itemize}
These conditions are aggregated using learnable convolution layers. 
Finally, the condition maps of all objects are averaged to form the control signal.

To control the diffusion model, we add the condition map to the input of its denoising network. 
The model is then fine-tuned on the dataset described in \cref{chap:dataset}. 
During inference, we further enhance control by manipulating cross-attention layers of the denoising network.
Specifically, for each visual concept, we restrict the attention map of image patches inside its bounding box, giving priority to the words of its text description.

\subsection{Interactive Editing with Panel Manipulation}
\label{sec:method_editing}
The image generation process of \method allows users to access the semantic panel for further image editing. 
Unlike complex and non-intuitive prompt engineering, editing images using \method is more natural and straightforward. 
Each editing operation corresponds to the update of visual concepts within the semantic panel. 
Considering the structure of semantic panel, we define the following six \textbf{unit operations}: 
1) \textit{adding} new objects, 2) \textit{removing} existing ones, 3) \textit{replacing} one with something, 4) \textit{resizing} objects, 5) \textit{moving} objects, and 6) \textit{re-editing} the attributes of objects.
Users can perform these operations manually or rely on the assistance of an LLM. 
For example, ``\textit{moving the ball to the left}'' can be achieved through a graphical user interface using drag-and-drop, or through an instruction-based chatting procedure with the help of an LLM.
We could also continuously update the semantic panel to refine the image progressively, resulting in more accurate and personalized outputs.

After updating the semantic panel, new visual concepts are utilized to generate the edited image latent.
To avoid unnecessary alterations to the original image, we confine the edits to editable regions using a binary mask $\mathbf{M}_e$.
Based on the difference between previous and new semantic panel, it is easy to determine the editable areas, \textit{i.e.} bounding boxes of adjusted visual concepts.
Assuming the latent representation of the original and current denoising process as $\mathbf{x}_t^{old}$ and $\mathbf{x}_t^{new}$, then the updated representation becomes $\mathbf{\hat{x}}_t^{new} = \mathbf{M}_e \mathbf{x}_t^{new} + (1 - \mathbf{M}_e) \mathbf{x}_t^{old}$.

\subsection{Semantic Panel Dataset}
\label{chap:dataset}
To support efficient training of \method, we build up a fully-automatic pipeline for preparing datasets, consisting of \textit{attribute extraction} and \textit{dataset augmentation}.

\vspace{1mm}
\noindent
\textbf{Attribute Extraction}.
We first collect a large set of 50M image-text pairs from multiple resources, \eg LAION~\cite{schuhmann2022laion} and WebVision~\cite{webvision}. 
For each image-text pair, attributes of all the visual concepts are extracted in the following order:


\begin{figure}[t]    
  \centering
   \includegraphics[width=0.95\linewidth]{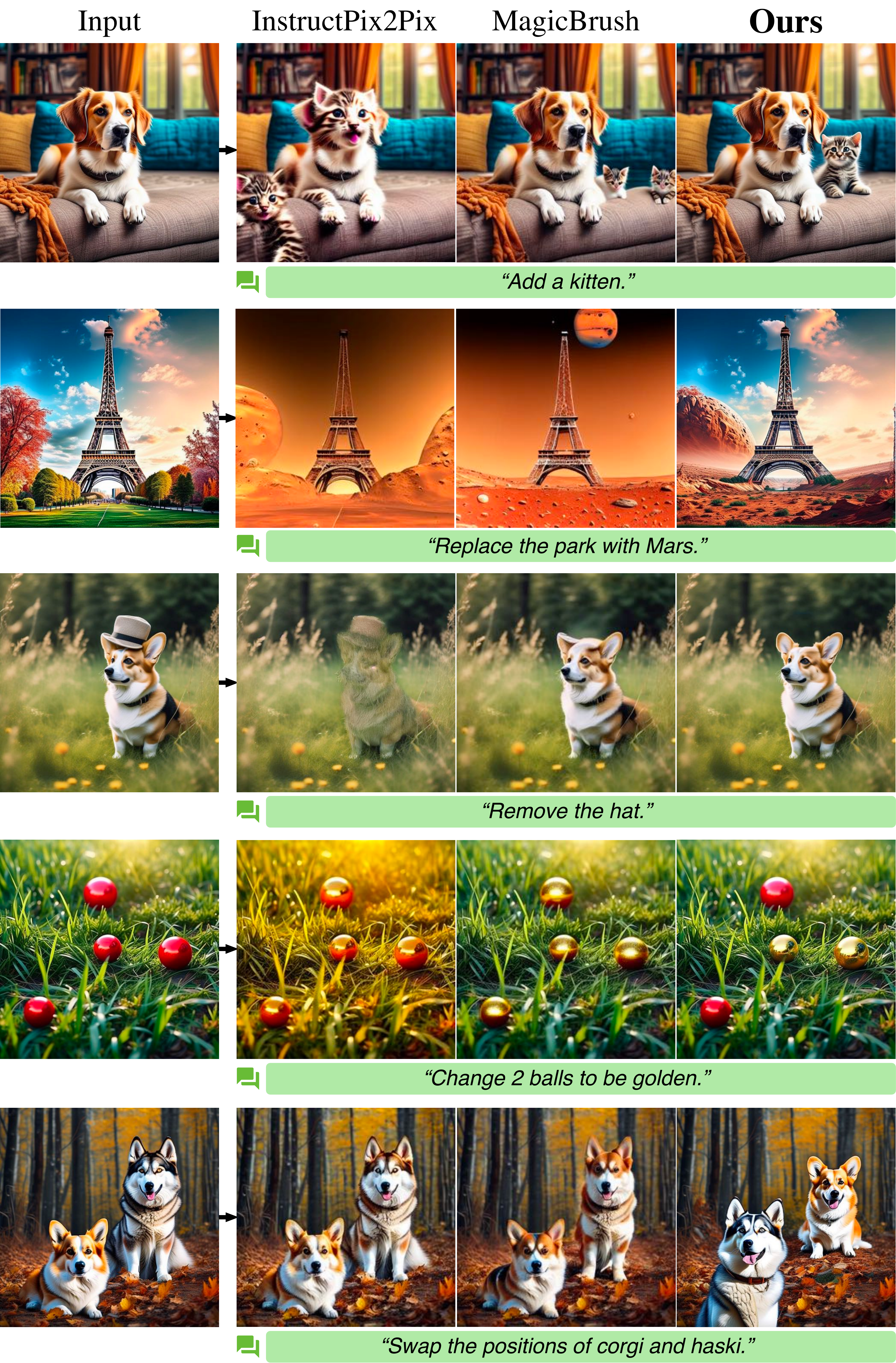}
    \vspace{-2mm}
   \caption{
   \textbf{Comparison on instruction editing} between \method and representative methods, using unit operation prompts.
   }
   \vspace{-6mm}
   \label{fig:edit_comp}
\end{figure}

\begin{figure*}[!t]    
  \centering
   \includegraphics[width=0.95\linewidth]{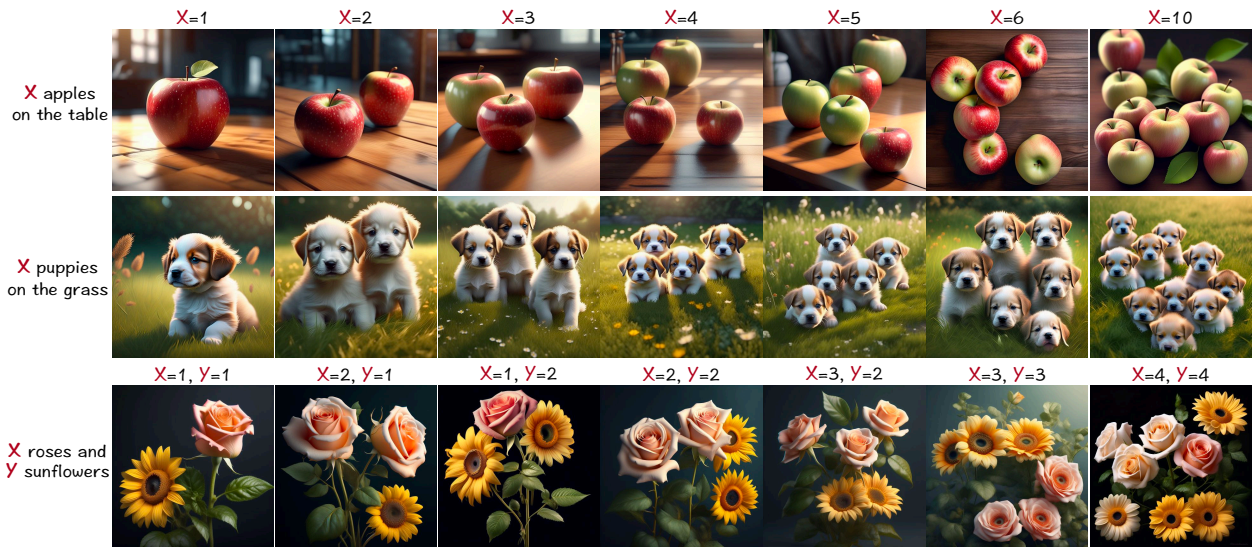}
    \vspace{-1mm}
   \caption{
Samples generated by \method on \textbf{quantity-awareness} prompts. }
   \label{fig:counting}
    \vspace{-2mm}
\end{figure*}
\begin{figure*}[!t]     
  \centering
   \includegraphics[width=0.95\linewidth]{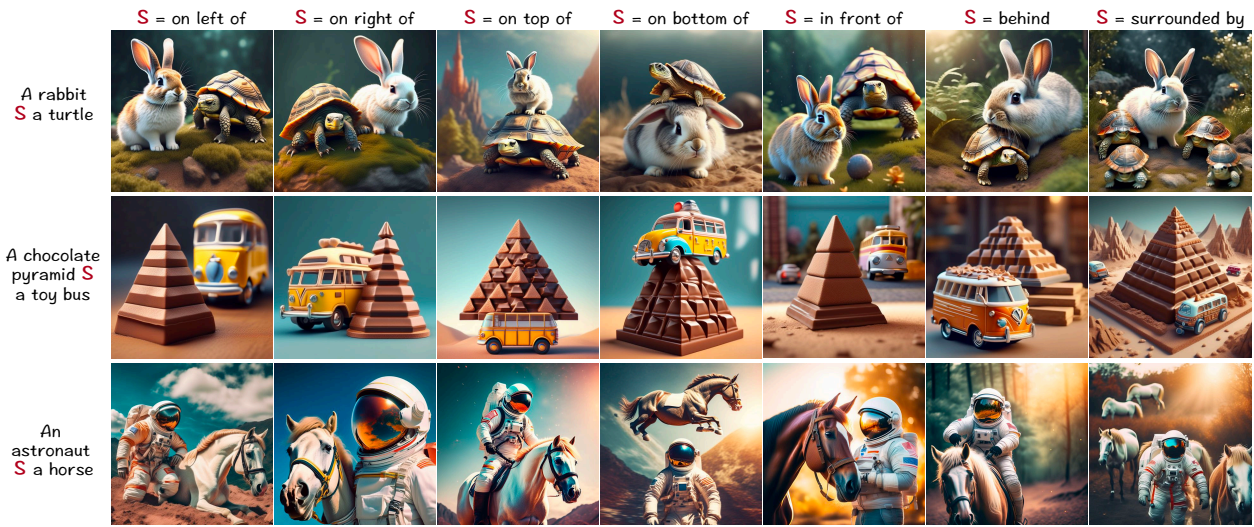}
    \vspace{-2mm}
   \caption{
   Samples generated by \method on \textbf{spatial relationship} prompts.
   }
   \vspace{-4mm}
   \label{fig:spatial}
\end{figure*}


\begin{figure*}[!t]    
  \centering
   \includegraphics[width=\linewidth]{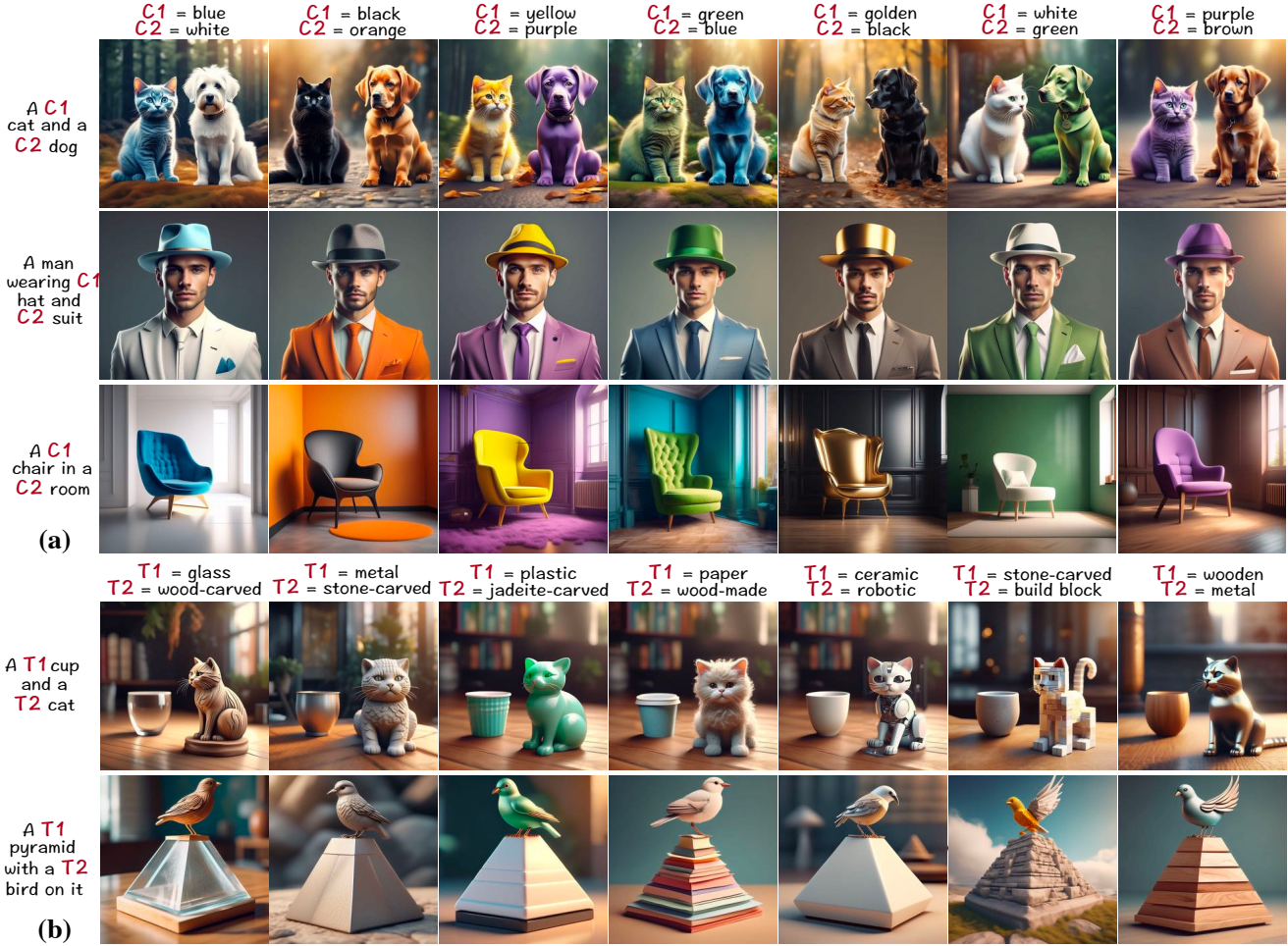}
    \vspace{-6mm}
   \caption{
  Samples generate by \method on \textbf{attribute binding} prompts, including the (a) color binding and (b) texture binding. For clear comparison, the random seed is fixed to preserve the spatial arrangement in one row.
   }
   \vspace{-1mm}
   \label{fig:attr_bind}
\end{figure*}

\begin{figure*}[!t]    
  \centering
   \includegraphics[width=\linewidth]{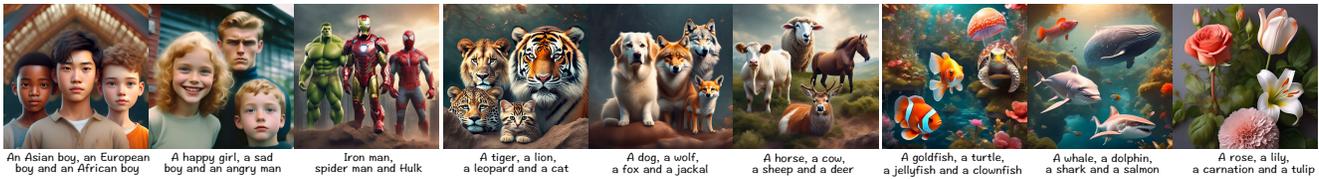}
    \vspace{-6mm}
   \caption{
   Samples generated by \method on \textbf{multi-object} prompts.
   }
       \vspace{-2mm}
   \label{fig:multi_ins}
\end{figure*}


\noindent \textit{(i) Description and Box:} 
Grounding DINO~\cite{groundingdino} is used to extract a list of objects with text descriptions and bounding boxes.
We then filter out meaningless descriptions, and remove highly-overlapped boxes with the same description.

\noindent \textit{(ii) Colors:} 
For each bounding box, we first use SAM~\cite{sam} to get its segmentation mask.
Each pixel inside the mask is mapped into the index of its closest color in a 156-colored palette.
We count the index frequency, and pick the top-6 colors with proportions larger than 5\%.

\noindent \textit{(iii) Keypoints:}
The keypoints are sampled within the SAM~\cite{sam} mask using the FPS algorithm~\cite{pointnet2}.
Eight points are sampled, with an early stopping when the farthest distance of FPS reaches a small threshold of $0.1$.

\vspace{1mm}
\noindent
\textbf{Dataset Augmentation.}
We empirically find it efficient to augment the dataset with following strategies:

\noindent \textit{(i) Synthesised Captions:}
The original caption of an image might ignore some objects, resulting in incomplete semantic panels. 
To address this issue, we utilize LLaVA~\cite{llava} to find out images with multiple objects and generate more detailed captions for them.

\noindent \textit{(ii) Mixing Pseudo Data:}
To enhance the ability of spatial arrangement, we create pseudo samples using manual rules.
We generate random prompts from a pool of objects with varying orientations, colors, and numbers.
Next, we synthesize the semantic panel by arranging them randomly according to specified rules.

\section{Experiments}
\label{sec:exp}

\subsection{Experimental Setup}
For the \textbf{text-to-panel} task, we select the open-sourced Llama-2~\cite{llama2} 13B version as our LLM.
To enable attribute generation for each parsed object, we fine-tune the LLM with LoRA~\cite{lora} for 10K steps with a batch size of 64.
The final optimized module for each attribute generation task contains 6.25M parameters, making it easy to switch between tasks.
The datasets are sampled with a mixture of 50\% probability from subsets with raw captions, 45\% from synthesized captions, and 5\% from pseudo data.

\noindent
For the \textbf{panel-to-image} task, we fine-tune a pre-trained latent diffusion model with 3B parameters on our constructed dataset with visual concepts. 
The fine-tune process contains 40K steps with a batch size of 128.
Training samples are evenly distributed between raw and synthesised captions.
To prioritize attribute conditions over text conditions in the model, we apply a drop rate of 0.7 to the text conditioning.

\subsection{Evaluation on Text-to-Image Alignment}

\textbf{Qualitative Evaluation.}
We expect \method to generate images that align better with text.
In this section, we examine its alignment ability with various types of prompts, which are known to be challenging for existing methods:

\begin{figure*}[!t]    
  \centering
   \includegraphics[width=0.97\linewidth]{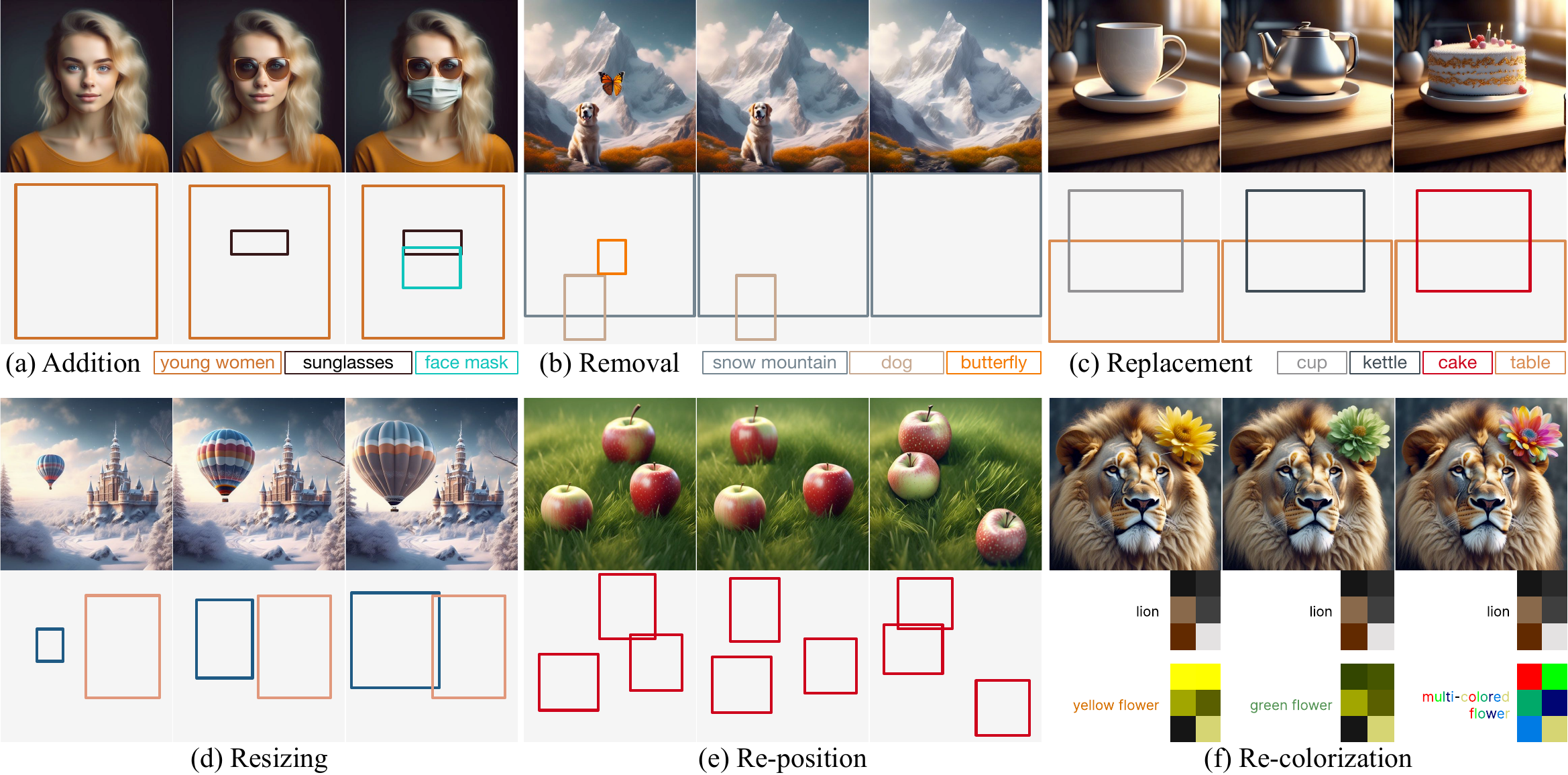}
    \vspace{-3mm}
   \caption{The editing results and corresponding panel update for each \textbf{unit operation}.}
       \vspace{-3mm}
   \label{fig:unit_ops}
\end{figure*}


\noindent
\textit{Quantity}: 
Existing model struggles to generate objects in the exact requested number. 
\cref{fig:counting} shows that \method is more sensitive to the varying numbers of objects.

\noindent
\textit{Spatial Relationship}:
We examine the spatial awareness of \method with 7 types of relationships in \cref{fig:spatial}.
Results show its ability to properly arrange object positions.

\noindent
\textit{Attribute Binding}: 
We show cases in \cref{fig:attr_bind}  on objects with varying attributes.
\method explicitly distinguishes different objects, thus allowing for precise assignment of attributes to each object without any cross-influence.

\noindent
\textit{Multiple Objects}:
When generating multiple objects with a similar appearance, existing models might confuse them together.
\cref{fig:multi_ins} shows that \method successfully generates groups containing similar people, animals or plants.

\vspace{2mm}
\noindent
\textbf{Quantitative Evaluation.}
We further evaluate the alignment with quantitative metrics.
For attribute binding and spatial relationship, we use the validation prompts and metrics from T2I-CompBench~\cite{t2icompbench}, with 300 prompts for each subset.
For quantity, we generate 300 prompts containing 1 to 5 objects in same type.
We set the score as the proportion of results that generate correct number of objects.
For multiple objects, we start by collecting 30 groups, each containing 4 similar objects such as tiger, lion, cat, and leopard.
Next, for each group, we generate 10 prompts that include 2 to 4 different objects.
The metric used is the BLIP-VQA score~\cite{t2icompbench}.

\cref{tab:benchmark} shows the evaluation results. \method outperforms existing methods, including end-to-end models and inference-optimized strategies.
In particular, it shows great improvement on the spatial relationship and quantity-awareness tasks.
We also compare with our pre-trained base model.
The improvement suggests that \method could enhance the prompt following of an existing model with semantic panel control.

\noindent
\textbf{Visualized Comparison.}
To compare with existing models, we visualize the results on different types of prompt in \cref{fig:gen_comp}.
We compare \method with LLM-grounded diffusion (LMD)~\cite{lian2023llmgroundeddiffusion}, Stable Diffusion XL 1.0~\cite{sdxl}, DALL·E 3~\cite{dalle3}, and Midjourney~\cite{midjourney}.
\method achieves competitive performance in prompt following, while maintaining the fidelity of its generation.
It is noteworthy that \method demonstrates improved alignment in terms of quantity-awareness and spatial relationship, which is consistent with the quantitative results in \cref{tab:benchmark}.


\begin{table}[!t]
\setlength\tabcolsep{2pt}
\def\w{20pt} 
\caption{%
    \textbf{Quantitative results for alignment assessment} on various benchmarking subsets. The best and second results for each column are \textbf{bold} and \underline{underlined}, respectively.
}
\vspace{-8pt}
\centering\footnotesize
\begin{tabular}{l@{\extracolsep{2pt}}c@{\extracolsep{4pt}}c@{\extracolsep{4pt}}c@{\extracolsep{3pt}}c@{\extracolsep{3pt}}c@{\extracolsep{2pt}}c}
\shline
\multirow{2}{*}{\textbf{Method}}             & \multicolumn{3}{c}{\textbf{Attribute Binding}}                                               & \multirow{2}{*}{\textbf{Spatial}} & \multirow{2}{*}{\textbf{Quantity}} & \multirow{2}{*}{\textbf{Multi-obj.}} \\
\cline{2-4} & Color & Texture & Shape & & &                         \\
    \shline
SD v1.5~\cite{stablediffusion2}                             & 0.3730                    & 0.4219                      & 0.3646                    & 0.1312                      & 0.1801                                        & 0.4255                                          \\
SD v2.1~\cite{stablediffusion2}                             & 0.5694                    & 0.4982                      & 0.4495                    & 0.1738                      & \underline{0.2337}                                        & 0.5562                                          \\
Composable~\cite{composable_diffusion} & 0.4063 & 0.3645 & 0.3299 & 0.0800 & - & - \\
Structured~\cite{structured_diffusion} & 0.4990 & 0.4900 & 0.4218 & 0.1386 & - & -  \\
Attn-Exct~\cite{attn_exct} & 0.6400 & 0.5963 & 0.4517 & 0.1455 & - & - \\
GORS~\cite{t2icompbench} & \underline{0.6603} & \underline{0.6287} & 0.4785 & 0.1815 & - & - \\
SDXL (b2r)~\cite{sdxl}                            & 0.6050                    & 0.5446                      & 0.4780                    & 0.2086                                               & 0.1992                                       & 0.5905                                          \\
SDXL (bpr)~\cite{sdxl}                            & 0.6132                    & 0.5331                      & \underline{0.4896}                    & \underline{0.2097}                                                      & 0.1839                                        & \underline{0.6221}                                          \\[-0.6ex]
\midrule
\\[-3ex]
\textit{Base model} & 0.5446 & 0.5970 & 0.4732 & 0.1833 & 0.2337 & 0.5579 \\
\textbf{\method (Ours)}    & \textbf{0.6893}           & \textbf{0.6325}             & \textbf{0.4934}           & \textbf{0.3167}                      & \textbf{0.2720}                                        & \textbf{0.6400}     \\
\shline
\end{tabular}
\vspace{-4mm}
  \label{tab:benchmark}%
\end{table}%

\begin{figure*}    
  \centering
   \includegraphics[width=\linewidth]{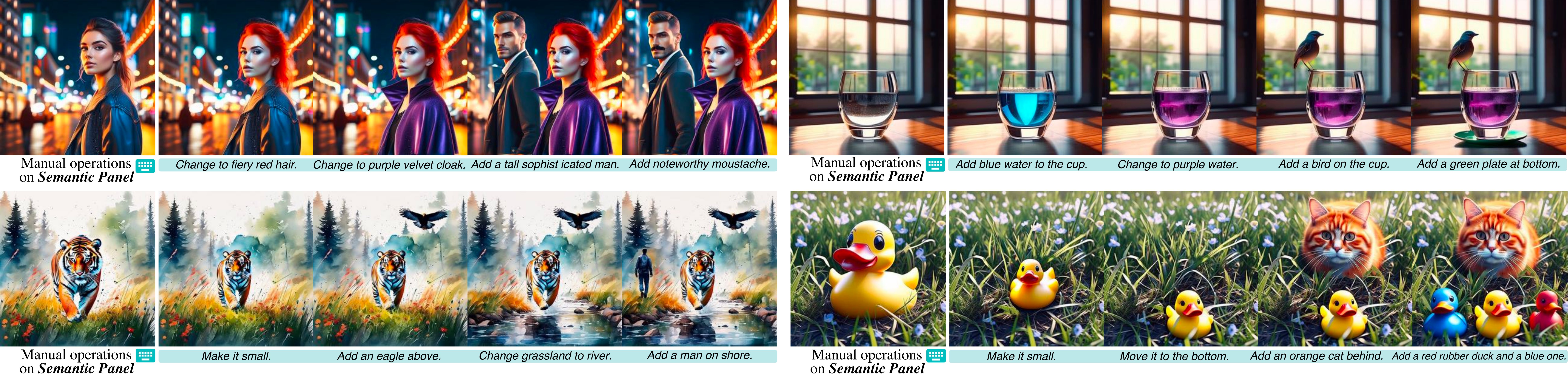}
    \vspace{-8mm}
   \caption{Results of 
   \textbf{continuous generation} with multi-round editing chains, consisting of unit operations.
   }
   \vspace{-1mm}
   \label{fig:compounded_edit}
\end{figure*}

\begin{figure*}    
  \centering
   \includegraphics[width=\linewidth]{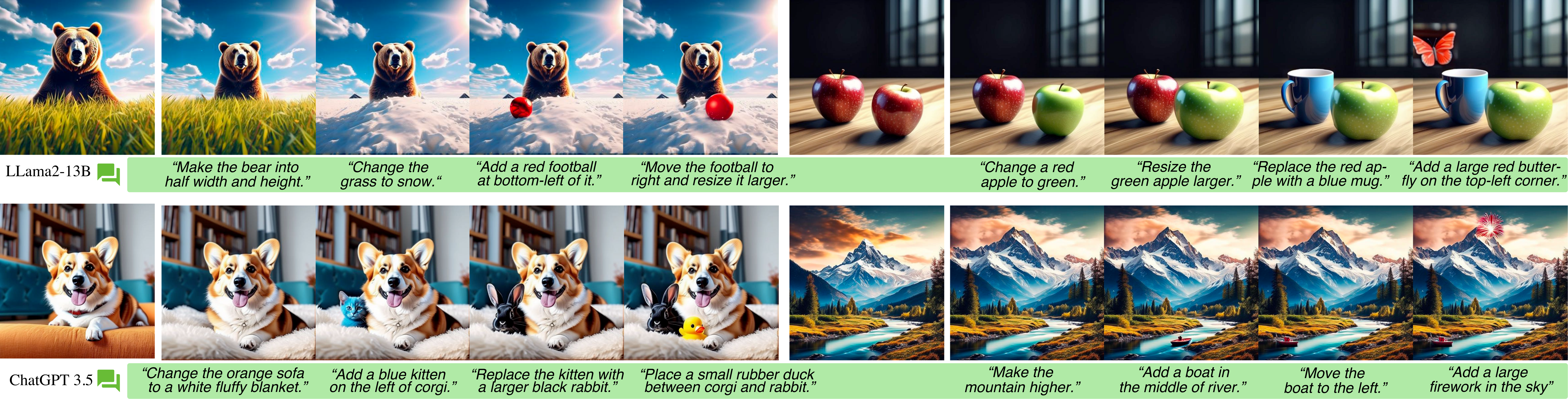}
    \vspace{-6mm}
   \caption{Results of 
   \textbf{chatting-based generation} in natural instructions with different LLMs . Refer to \cref{fig:teaser} for the results of ChatGPT-4.}
   \label{fig:chatting_edit}
\end{figure*}

\begin{figure*}[!t]     
  \centering
   \includegraphics[width=\linewidth]{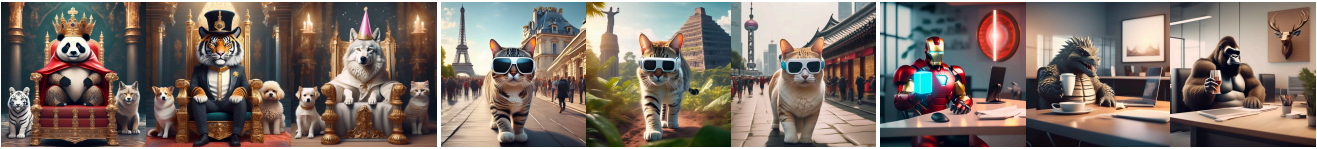}
    \vspace{-7mm}
   \caption{
   Samples generated by \method with \textbf{similar layouts}.
   }
   \vspace{-4mm}
   \label{fig:similar_layout}
\end{figure*}

\subsection{Evaluation on Interactive Generation}
Based on the generated image with its semantic panel, \method can make further edit to the image at a high semantic level.
We first evaluate \method's performance on unit editing operations.
Then we expand its capabilities to include multi-round editing with compounded operations.
Lastly, we enhance \method by incorporating the intelligence of LLMs to enable chatting-based editing.

\vspace{2mm}
\noindent
\textbf{Unit Operation} defined in \cref{sec:method_editing} is the basis for all editing operations.
Most editing intentions can be considered as one or a combination of the unit operations.
\cref{fig:unit_ops} shows the correspondence between each unit operation and the update of semantic panel.
We compare the editing ability \wrt unit operations with Instruct-Pix2Pix~\cite{instruct_p2p} and MagicBrush~\cite{magic_brush} in \cref{fig:edit_comp}.
We can see that \method could better preserve the non-editing area, and achieve more flexible operations, \eg swapping positions.

\vspace{2mm}
\noindent
\textbf{Compounded Operations.}
Based on the unit operations, we further apply \method for continuous editing with compounded operations.
In \cref{fig:compounded_edit}, we present examples of progressively creating images with complex scenes.
During this interactive creation process, users can refine the image step-by-step by replacing unsatisfying objects, adding more details, and experimenting with various attributes.
The interactive nature also enables additional applications, such as generating images with similar layouts, as shown in \cref{fig:similar_layout}.
To achieve this, we first generate an image as the base and then sequentially replace objects in it.

\vspace{2mm}
\noindent
\textbf{Chatting-based Editing.}
We also use LLMs to automatically map editing instructions into updates of the semantic panel.
To accomplish this, we introduce new system prompts that are specifically designed for this task. These system prompts request the LLM to understand the current semantic panel and the editing instruction, and then generate the updated panel.
Please refer to the Supplementary Material for more details.
\cref{fig:chatting_edit} and \cref{fig:teaser} (c) present cases using Llama2-13B~\cite{llama2}, ChatGPT-3.5~\cite{chatgpt}, ChatGPT-4~\cite{gpt4}, respectively.
During the evaluation, we observed that LLM has the ability to understand more natural instructions. 
For instance, the instruction "\textit{the mushroom is eaten}" indicates the need to remove it, while "\textit{the mushroom grows higher}" implies increasing its height while keeping its bottom position intact. 
The results demonstrate the potential of \method as a unified image creation system that supports sequential instructions with chatting.

\section{Conclusion}

In this paper, we present \method, a new approach that tames existing diffusion models to better follow the painting and editing instructions.
The semantic panel in \method is introduced as a generative middleware between text and image. 
It helps relieve the pressure of directly mapping complex prompt to image.
The panel is firstly constructed using visual concepts parsed by LLM from the given prompt.
It then serves as control signal to complement the generation of diffusion models.
\method follows painting instruction without ignoring detailed description of each concept in prompt.
Furthermore, by adjusting the semantic panel with manual or LLM-based operations, \method enables interactive editing of previous generated images.
We demonstrates that with the fully automatic control of LLM, \method shows potential as a flexible chat-based image creation system, where any existing diffusion model can be incorporated as the generator for interactive generation.

\break

\clearpage
\maketitlesupplementary
\renewcommand\thefigure{A\arabic{figure}}
\renewcommand\thetable{A\arabic{table}}
\renewcommand\theequation{A\arabic{equation}}
\setcounter{page}{1}
\setcounter{equation}{0}
\setcounter{table}{0}
\setcounter{figure}{0}
\appendix

\section{Illustration of the Complete Workflow}

\begin{figure}[ht]    
  \centering
   \includegraphics[width=0.93\linewidth]{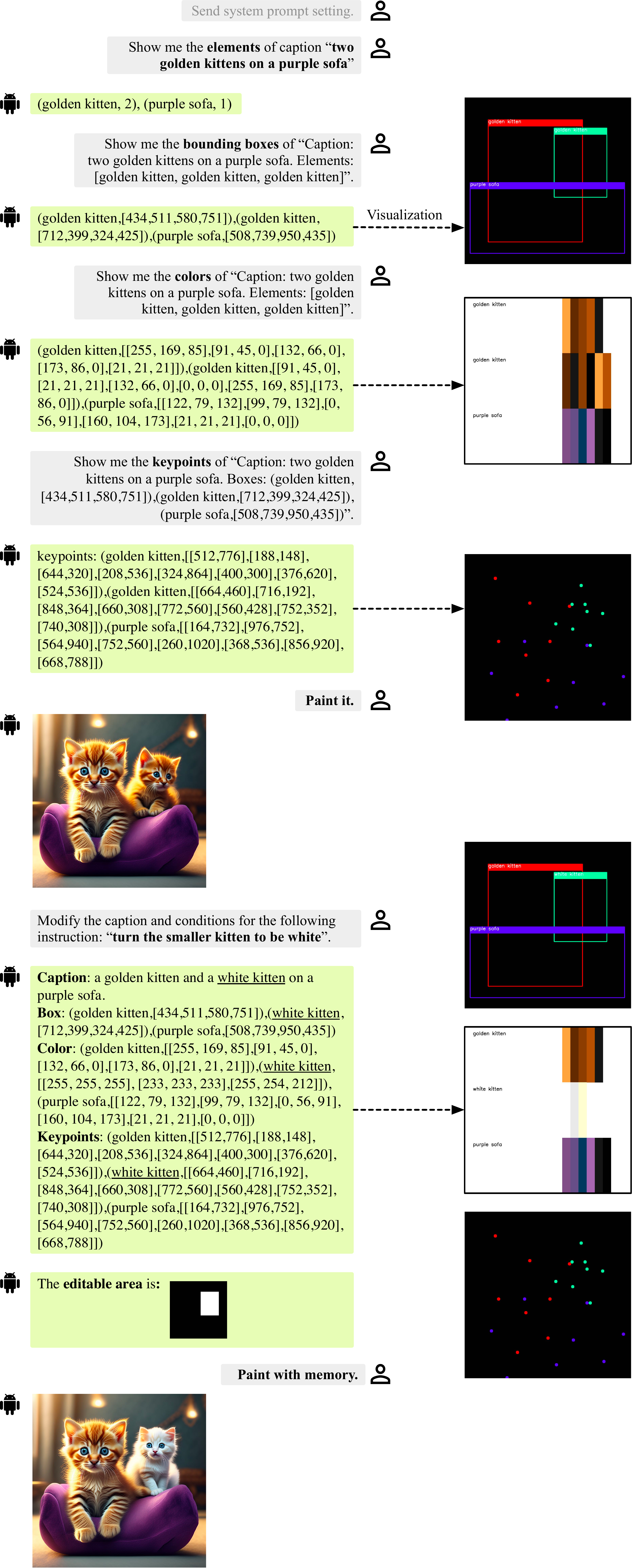}
   \caption{Examples of \textbf{a complete generation process} by \method with painting and editing instructions. }
   \vspace{-6mm}
   \label{fig:sup_example_chat}
\end{figure}

In this section, we provide a complete example of the workflow, including painting and editing instructions.
\cref{fig:sup_example_chat} illustrates the conversation process for requesting LLM to create and manipulate the semantic panel, with step-by-step instructions.
Visualized internal results and images are included for better understanding.
Based on the explicit design of the semantic panel, \method presents a fully-automatic pipeline for assigning and manipulating images using a conversational approach.

\section{Dataset Construction}
In this section, we present the details of our data preparation pipeline, including the attribute extraction and dataset augmentation process.
Furthermore, we showcase visualizations of samples from various parts of the semantic panel dataset.


\subsection{Attribute Extraction}
\textbf{Description and Box.}
Given an image with a complete caption, we use Grounding DINO~\cite{groundingdino} to detect all the visible object boxes along with their corresponding descriptions in the caption.
After the inference of Grounding DINO, we filter out redundant boxes that have general or meaningless descriptions, \eg ``an image'', ``objects'', \etc.
We also remove boxes that have the same description as another box, with an IoU larger than $0.9$.

\vspace{1mm}
\noindent
\textbf{Colors.}
For each object, we use SAM~\cite{sam} mask to extract all its pixels.
We first construct a color palette with CIELab~\cite{cielab} color space, which consists of 11 hue values, 5 saturation values and 5 light values.
We calculate the color index of each pixel by searching for its nearest RGB value in the palette.
Then we count the frequency of all color indices for the object, filter out indices with frequencies smaller than 5\%, and pick the top-6 indices as the color representation. 
The final output of color attribute is a set of high frequent indices. 

\vspace{1mm}
\noindent
\textbf{Keypoints.}
We use the farthest point sampling (FPS) algorithm~\cite{pointnet2} to sample keypoints within the SAM mask. 
Specifically, we define the candidate set as all the pixel coordinates $(x,y)$ inside the mask area. 
We start by randomly selecting a point and adding it to the sampled set. 
Then, for each iteration, we choose a point from the candidate set that has the farthest distance towards the sampled set. 
The distance from a point to a point set is determined by its distance to the closest point in that set. 
We stop sampling when the size of the sampled set exceeds $8$ or when the farthest distance is smaller than $0.1$ (the distance is normalized to the range $[0, 1]$).

\begin{figure*}[htbp]    
  \centering
   \includegraphics[width=0.98\linewidth]{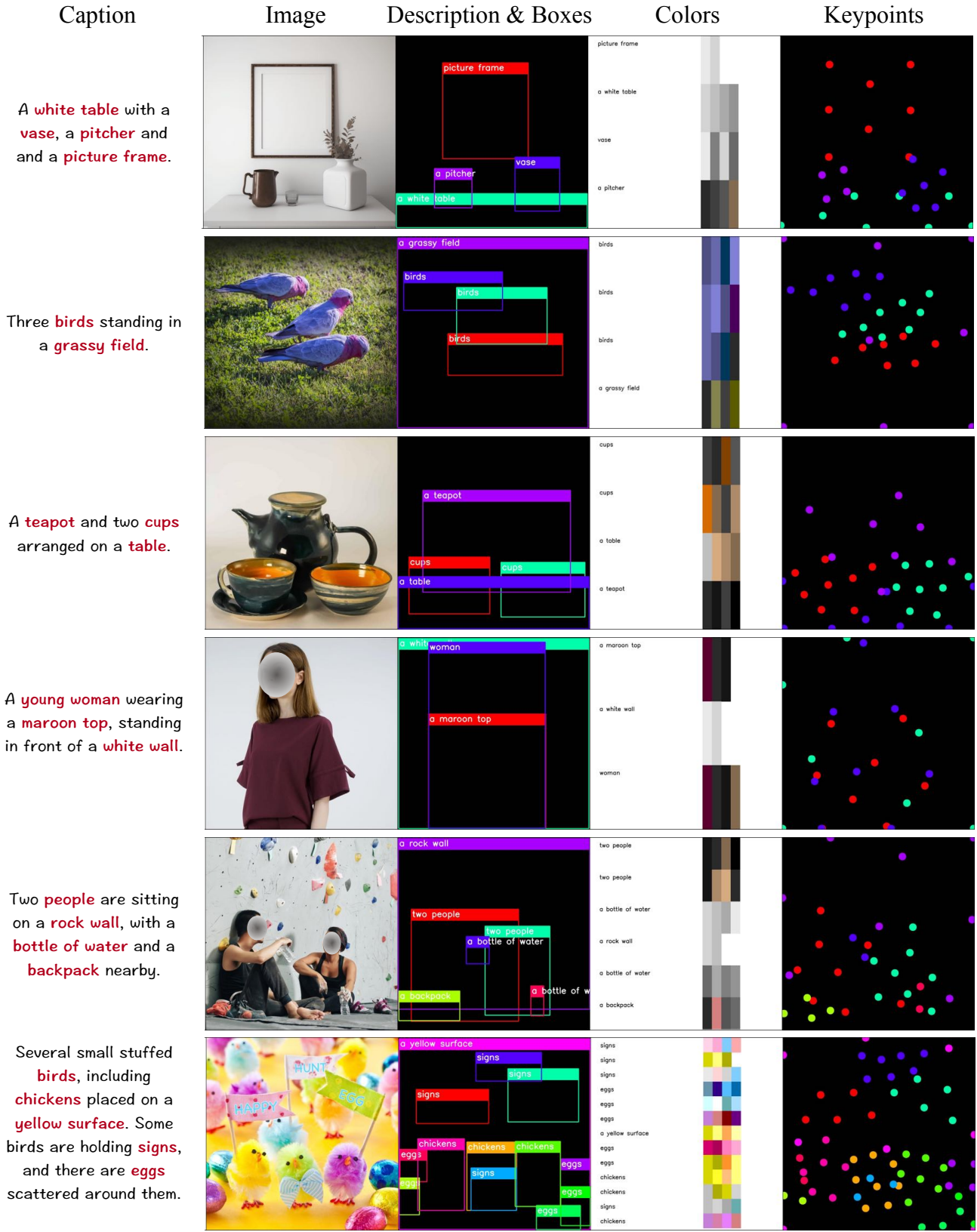}
    \vspace{0mm}
   \caption{\textbf{Visualization of samples} in the semantic panel dataset, with all the extracted attributes based on the orignal text-image pair.}
   \label{fig:supp_dataset_vis}
\end{figure*}

In \cref{fig:supp_dataset_vis}, we show visualizations of samples with all extracted attributes.
Since all the attributes in semantic panel can be automatically extracted for existing image-text pairs, we can easily scale up the dataset and enable efficient training of \method.


\subsection{Dataset Augmentation}
\textbf{Generate Synthesised Captions.}
We use Llava~\cite{llava} to generate captions for dataset augmentation.
Llava is a visual question answering model that provides answers to questions based on the content within a given image.
First, we ask it to pick out images with only one object by asking: \textit{``Is there only one element or object in the image?''} 
Images with the answer \textit{``No''} are kept because their raw captions usually overlook the details of some objects.
Next, we request captions for these images with a limited length: "Analyze the image in less than twenty words, focusing on all objects in this image."

\vspace{1mm}
\noindent
\textbf{Generate Pseudo Data.}
The pseudo samples are generated by creating random prompts and arranging random visual elements for them according to specific rules.
Firstly, we generate random prompts from a pool of diverse objects, and assign varying colors and numbers to each prompt.
In cases involving spatial relationships, we randomly specify the relative positions of the objects.
For the spatial arrangement of their bounding boxes, we create a large set of prompts and randomly assign positions. 
We then select appropriate samples based on the criterion of maximizing the separation of elements. 
This effectively prevents the issue of object concentration in pseudo data.
The spatial arrangement is specifically designed for prompts with spatial relationships, such as \textit{``on the left of''}.


\begin{figure*}[htbp]    
  \centering
   \includegraphics[width=\linewidth]{figure/supp/supp_system_prompt.pdf}
    \vspace{-2mm}
   \caption{\textbf{System prompts} for all the LLM-based tasks in \method. We leave ``\{\}'' in red for positions of depended conditions.}
   \label{fig:supp_system_prompt}
\end{figure*}


\section{Implementation Details of Text-to-Panel}

In this section, we show the details of LLM-based text-to-panel generation in \method.
This process is conducted step-by-step for different attributes in panel.
For all the attribute generation, we carefully search for a system prompt to leverage the zero-shot ability of LLM. All the system prompt templates are shown in \cref{fig:supp_system_prompt}. 

\subsection{Description Generation}
The description generation is the first step, which finds out all the elements to be appeared in the image.
The task is a pure language-based problem, without requiring knowledge on visual space.
Therefore, we directly leverage the zero-shot ability of LLM for this task.
As the system prompt shown in \cref{fig:supp_system_prompt}, we define a specific output format on this task.
Instead of a raw set of element descriptions, we request LLM to generate each unique element with its number, \textit{e.g., ``(cat, 3)''}.
We empirically observe that such a strategy results in better performance of description prediction, especially for objects with larger number of amount.
Furthermore, we also request the LLM to ignore style descriptions and all invisible objects.
We find it works well to ignore the \textit{unwanted} objects, such as \textit{``a sky without cloud''}.

\subsection{Box Generation}
It is more challenging to generate bounding boxes of predicted elements in the above process.
The region information of bounding boxes gets closer to the image modality, requiring more knowledge on spatial distribution.
First, we design the system prompt to teach the LLM understanding the coordinate system of image, \textit{i.e.} the x and y axis, with values increasing from left-to-right and top-to-bottom, respectively.
For the output format of bounding boxes, we find it useful to define it as $[x_c, y_c, w, h]$, where $(x_c, y_c)$ is the center point of the box, and $(w, h)$ is the width and height of box.
Different from the most commonly used $[x_1, y_1, x_2, y_2]$ indicating the top-left and bottom-right of the box, our used format is more friendly to LLM.
The size of box is fixed when moving to different position, which helps LLM to learn the relationship between object description and its size.

\subsection{Color Generation}
We define the color representation as discrete indices in a 156-colored palette.
Such a discrete representation is necessary to relax the range of output.
In practical, we test different strategies for color prediction:
(1) Use the name of each color in the palette. It can relax the color prediction as a easier task of language model, but restrict the size of palette for accurate representation.
(2) Use the color indices, and predict the index list. This strategy is hard to learn for LLM, without any knowledge on the color indices and their relationship.
(3) Use the color indices, and predict the RGB value list. This strategy helps LLM to understand the colors and relationships.
We choose the final strategy in our method.

\subsection{Keypoints Generation}
The keypoint generation is the most difficult task of LLM in \method.
We find that it is hard to prompt LLM with a good initial prediction, by carefully setting system prompt.
Thus, we focus on helping LLM to output in a correct format, and learn the ability of keypoint prediction in the fine-tune stage.
As shown in \cref{fig:supp_system_prompt}, we also provide the predicted box of each object, which restrict the distribution of keypoints to be inside the box.

On an NVIDIA A100 GPU, the averaged runtime is $6.75 \pm 1.65$s for text-to-panel with Llama2-13B.

\section{Implementation Details of Panel-to-Image}

In this section, we present the details of panel-to-image, a controllable image synthesis process based on predicted semantic panel.
As we have mentioned, the controlling strategy for panel-to-image contains two parts, \textit{i.e.} panel conditioning and 
attention restriction.

\subsection{Panel Conditioning}
We first encode each attribute in the semantic panel into a comprehensive condition:

(1) For the text description, we get its CLIP text embedding~\cite{clip} individually. We use the same CLIP weights as the main text-to-image model, but take the global sentence embedding instead of word embedding.

(2) For the bounding box, we draw a binary mask in the same shape of image latent.
The coordinates of box are resized into the latent space, \textit{i.e.,} $1/8$ of original value.
Then we set all positions inside the box as value $1$ in the mask.

(3) For the colors, we have got a list of color indices. Then we set a binary vector in size of 156 (same as the palette size), and set $1$ for the given color indices. The vector is then mapped to a feature vector with learnable linear projection.

(4) For the keypoints, we draw a binary mask same as box.
For each point, we draw a circle with radius of $6$, and set $1$ inside the circle.

All the conditions are then mapped by learnable convolutions into the same channel.
To merge the the conditions in different shape, we further repeat the 1D conditions (text and color) into the same shape of image, and multiply it with the binary mask of bounding box. 
Finally, we sum all conditions up, and average it over all objects.

Except for the training strategy in main paper, we also study the strategy based on ControlNet~\cite{zhang2023adding}, which digests the condition with a copied bypass encoder. 
Since the final condition map is a feature map in same shape of image latent, we can easily train a ControlNet for this task.
In practical, we find comparable performance for the two different strategy, and choose the previous one for efficiency.
To accompany \method with existing models, \textit{e.g.,} Stable Diffusion~\cite{stablediffusion2}, it would be better to choose ControlNet as a plug-in module for the base model.

\subsection{Attention Restriction}
In the above process, we use sentence embedding for semantic description.
When the phrase of description comes longer, \textit{e.g.} ``a red metal apple'', the generated image may loss some semantics.
To address this issue, we further introduce a controlling strategy for better alignment.
It works via rectifying the cross-attention layer of diffusion model.

The present diffusion model involves cross-attention between $N_I$ image patches and $N_T$ words of input prompt.
Given the generated semantic panel, we have already known the exact correspondence between patches and words.
Then our rectifying is to restrict the attention map to follow such correspondence.
We generate a attention mask $M \in \mathbb{R}^{N_I \times N_T}$ for such rectifying.
For each object, we first locate the index range $[i_s, i_e]$ as its text description in the whole prompting text, then locate $[j_s, j_e]$ as the related image patches inside the bounding box. 
Then the attention mask is set as $M[i_s: i_e, j_s, j_e] = 1$, otherwise $0$.
We apply attention mask for all the cross-attention layers in the diffusion model.

The cross-attention rectifying significantly improves the alignment of semantics. 
But it can not restrict the object to be located inside the box.
Thus, we combine it with the training-based panel conditioning together, and achieve better controlled generation.

On an NVIDIA A100 GPU, the averaged runtime is $19.28 \pm 0.19$s for panel-to-image with our pre-trained 3B UNet and 50 diffusion steps. 

\section{Failure Case Analysis}
\begin{figure}[hbp]    
  \centering
   \includegraphics[width=0.98\linewidth]{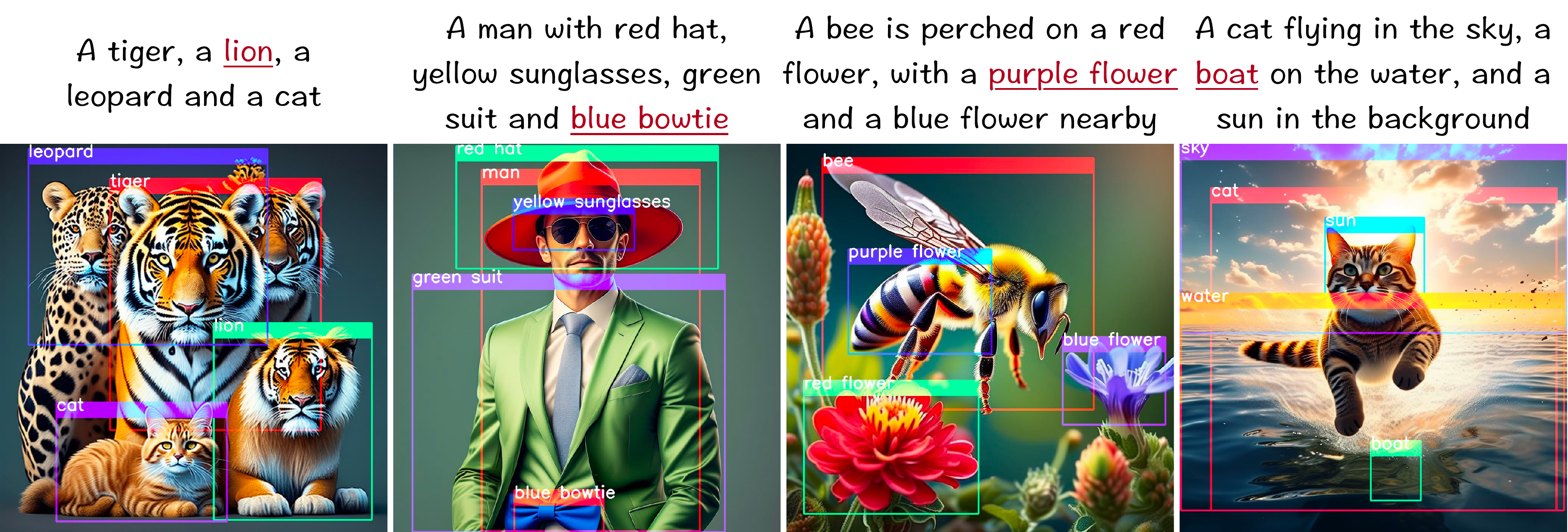}

   \caption{Failure cases of \method.}
      \label{fig:supp_failure_case}
    \
\end{figure}

We show failure cases in~\cref{fig:supp_failure_case}, including semantic confusion, wrong spatial relationship and missing objects.
The text-to-panel stage might generate results with wrong or highly-overlapped positions in such cases, leading to failed images in final generation.
As a preview, the users can refresh or adjust the elements before generating images for remedy.
It is also observed that the panel-to-image generation is not strictly controlled by the panel, and  shows some robustness to rectify improper layout from the first stage.

\section{More Results}
\begin{figure*}[htbp]    
  \centering
   \includegraphics[width=0.63\linewidth]{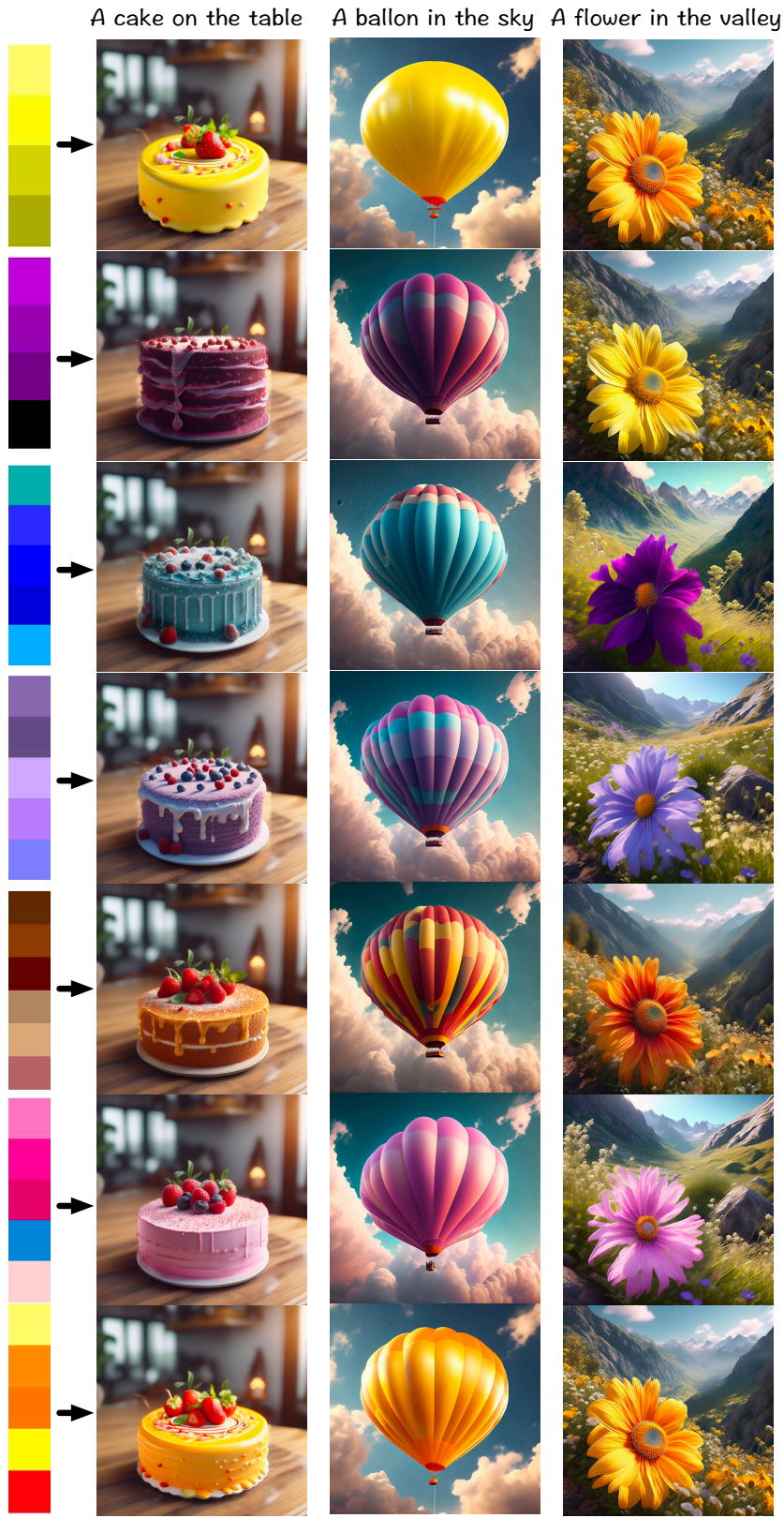}
    \vspace{0mm}
   \caption{Examples of \textbf{color editing}. }
   \label{fig:supp_edit_color}
\end{figure*}

\begin{figure*}[htbp]    
  \centering
   \includegraphics[width=0.75\linewidth]{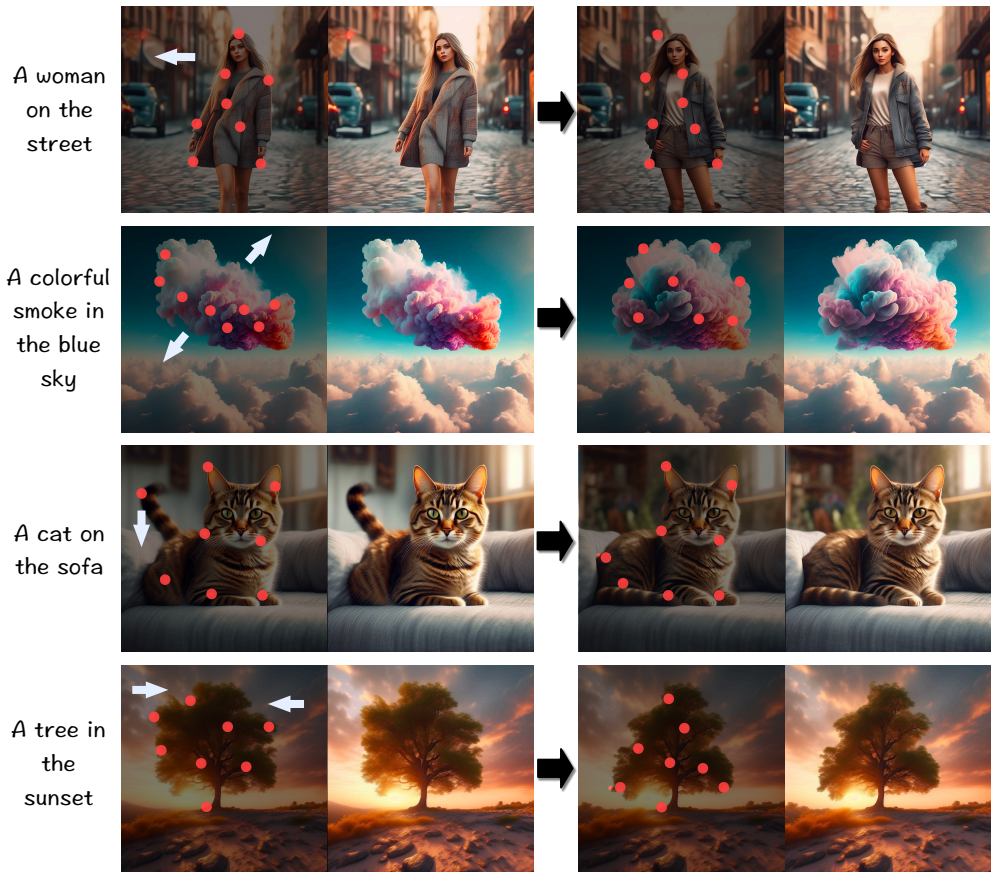}
   \caption{Examples of \textbf{shape editing}. The blue arrows indicate the direction of keypoints moving. }
   \label{fig:sup_edit_shape}
\end{figure*}

\begin{figure*}[!ht]    
  \centering
   \includegraphics[width=0.85\linewidth]{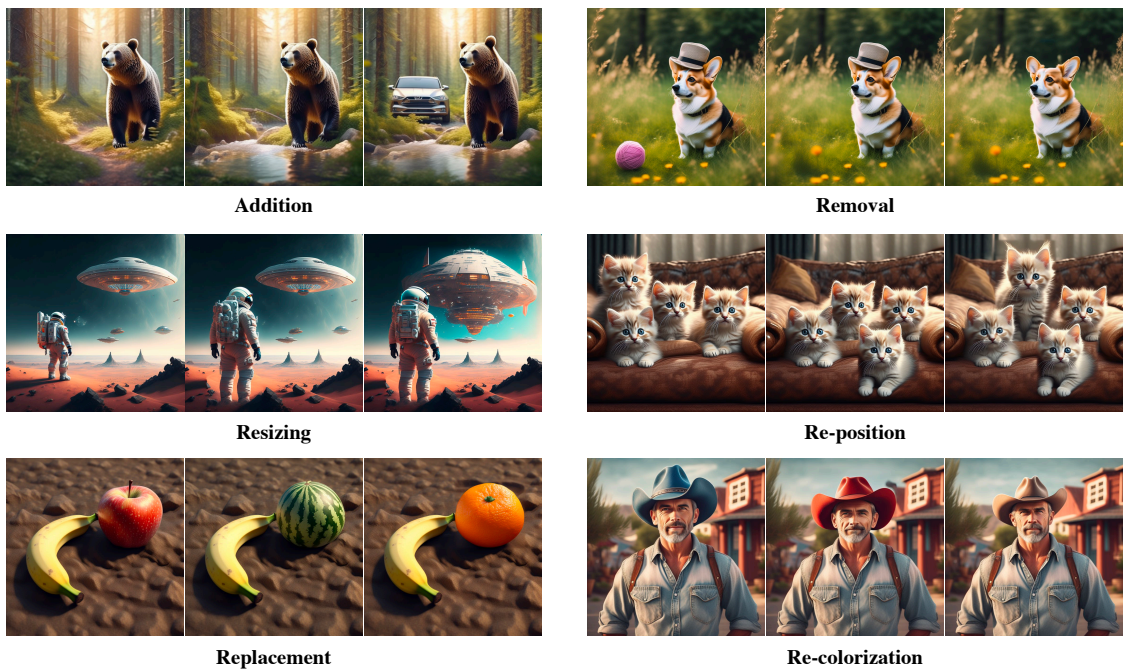}
   \caption{More editing cases of unit operations. }
   \label{fig:sup_more_edit}
\end{figure*}

In this section, we show more results of \method.
\cref{fig:supp_edit_color} shows editing samples with more detailed re-colorization by setting the color attribute.
\cref{fig:sup_edit_shape} shows shape-editing samples by re-arranging the keypoints, where we set a main direction to reshape the object.
\cref{fig:sup_more_edit} shows more editing results of the six unit operations.
\cref{fig:sup_more_res} shows more samples generated by \method.


\begin{figure*}[!ht]    
  \centering
   \includegraphics[width=\linewidth]{figure/supp/supp_more_res.pdf}
   \caption{More samples generated by \method. }
   \label{fig:sup_more_res}
\end{figure*}




{
\small
\bibliographystyle{ieeenat_fullname}
\bibliography{ref.bib}
}

\end{document}